\documentclass[letterpaper, 10 pt, conference]{ieeeconf}  % Comment this line out if you need a4paper

\IEEEoverridecommandlockouts                              % This command is only needed if 
\usepackage{graphics} % for pdf, bitmapped graphics files
\usepackage{epsfig} % for postscript graphics files
\usepackage{mathptmx} % assumes new font selection scheme installed
\usepackage{times} % assumes new font selection scheme installed
\usepackage{amsmath} % assumes amsmath package installed
\usepackage{amssymb}  % assumes amsmath package installed
\usepackage{amsmath,amssymb,amsfonts}
\usepackage{algorithmic}
\usepackage{graphicx}
\usepackage{textcomp}
\usepackage{xcolor}
\usepackage{booktabs}
\usepackage{array}
\usepackage{makecell}
\usepackage{tikz}
\let\labelindent\relax
\usepackage{enumitem}
\setlist[itemize]{leftmargin=1.2em, labelsep=0.6em, itemindent=0pt, listparindent=0pt}
\usepackage{multirow}
\usepackage{float}

\usepackage{authblk}

\title{\LARGE \bfseries ADM-Fusion: Adaptive Deep Multi-Sensor Fusion for Robust\\
Ego-Motion Estimation in Diverse Conditions}

\author{
Hasan Moughnieh$^{1}$,
Ibrahim Ghaddar$^{2}$,
Hadi Elham$^{1}$,
Imad H. Elhajj$^{1}$,
Daniel Asmar$^{2}$%
\thanks{$^{1}$Department of Electrical and Computer Engineering, American University of Beirut, Beirut, Lebanon.}%
\thanks{$^{2}$Department of Mechanical Engineering, American University of Beirut, Beirut, Lebanon.}
}

\begin{document}

% --- tighten float spacing safely (IEEE) ---
\setlength{\textfloatsep}{6pt}
\setlength{\intextsep}{6pt}
\setlength{\floatsep}{4pt}
\setlength{\abovecaptionskip}{2pt}
\setlength{\belowcaptionskip}{0pt}

\maketitle
\thispagestyle{empty}
\pagestyle{empty}

%%%%%%%%%%%%%%%%%%%%%%%%%%%%%%%%%%%%%%%%%%%%%%%%%%%%%%%%%%%%%%%%%%%%%%%%%%%%%%%%
\begin{abstract}
Robust multi-sensor fusion is essential for reliable autonomy in diverse and degraded environments where sensor reliability can fluctuate rapidly. Because different modalities fail in distinct ways, effective fusion should adaptively balance complementary cues rather than rely on fixed weighting. This adaptability is particularly important for ego-motion estimation, since accurate pose updates depend on the consistent integration of complementary sensor information. We propose ADM-Fusion, an end-to-end, deep learning–based multi-sensor fusion method designed to adapt to environmental changes and sensor degradation. ADM-Fusion employs an adaptive sensor mixture-of-experts framework with content-aware routing to dynamically assign weights to sensor inputs in real time. The system further incorporates separate translation and rotation branches, coupled via a cross-task attention mechanism to preserve task-specific specialization while enabling information sharing. ADM-Fusion is trained on the CARLA-Loc simulated dataset and subsequently fine-tuned on KITTI real-world data, demonstrating effective sim-to-real transfer. Experiments show that ADM-Fusion remains robust under degraded conditions competing with existing methods. 

% CHECK IF THERE IS A COMMON VRL GROUP ACCOUNT
% The code will publicly be available at \href{https://github.com/Hasanmog/ADM_Fusion}{https://github.com/Hasanmog/ADM_Fusion}

\end{abstract}

%%%%%%%%%%%%%%%%%%%%%%%%%%%%%%%%%%%%%%%%%%%%%%%%%%%%%%%%%%%%%%%%%%%%%%%%%%%%%%%%
\setlength{\parindent}{0pt}
\section{INTRODUCTION}
% outline :
% - Sensor Fusion In Autonomous Driving from general apps to odometry 
% - From classical to deep learning based sensor fusion 
% - Importance of robustness to environmental conditions
Ego-motion estimation is the process of estimating an agent’s three-dimensional position and orientation over time from onboard sensor data as it moves through an environment. This area represents an active field of research within the Computer Vision and Robotics communities. Accurate motion estimation is fundamental to many applications, including autonomous driving \cite{geiger2012kitti}, robotics \cite{7747236}, and virtual and augmented reality \cite{4538852}. This broad reliance underscores the need for continuous, accurate pose estimation in diverse and challenging environments.

Autonomous vehicles primarily rely on GPS for global positioning. In GPS-denied environments, these vehicles depend on onboard sensors to maintain accurate state awareness. Multi-sensor fusion fulfills this need by integrating complementary measurements from various modalities and dynamically adjusting their contributions as sensor quality fluctuates. However, under challenging environmental conditions, each sensor may experience distinct forms of degradation, such as visibility loss, signal attenuation, multi-path effects, or motion distortion. These factors cause the reliability of individual modalities to change over time, ultimately reducing overall fusion performance. Therefore, a robust multi-sensor fusion mechanism is needed to optimize each sensor's contribution based on the noise conditions.

Reliable sensor data is essential for effective fusion, yet each modality degrades under specific conditions. Cameras provide high-resolution imagery and rich semantic cues, but their performance deteriorates in low-light settings. LiDAR delivers accurate 3D geometry across a wide range of lighting conditions; however, heavy rain or fog can introduce significant noise and artifacts into its point clouds. Radar remains robust in adverse weather and supports long-range sensing and velocity estimation, but its lower spatial resolution limits fine-grained perception and classification. IMUs provide high-rate motion measurements that are critical for short-term odometry, yet their errors accumulate over time due to bias and drift.
\begin{figure}[t]
\centering
\includegraphics[width=\columnwidth,page=1]{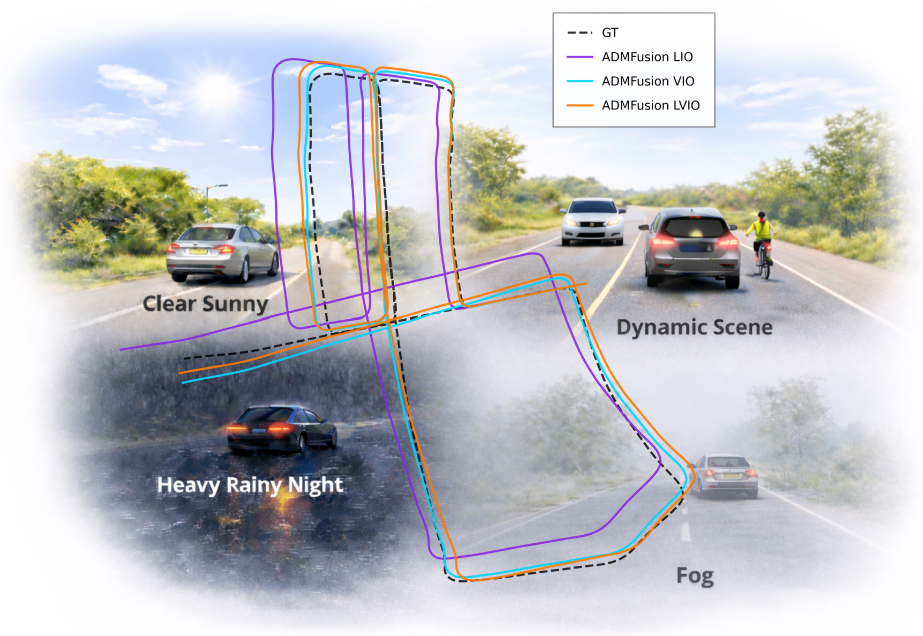} % page=1 means first page
\caption{Motivation for ADM-Fusion. Autonomous driving environments vary significantly across weather and dynamic conditions. By adaptively integrating multiple sensor modalities, ADM-Fusion maintains robust trajectory estimation across diverse scenarios.}
\label{fig:motivation}
\end{figure}
Recent advances in multi-sensor fusion for ego-motion estimation have progressed from geometry-centric pipelines that integrate modalities via hand-crafted measurement models and explicit optimization \cite{mur2017orb}, to hybrid approaches that incorporate learned perception modules while preserving model-based estimation objectives \cite{younes2024h}. More recently, end-to-end deep learning fusion models have emerged that predict motion directly from raw multi-modal streams \cite{javanmardghareshiran2021deeplio}. While these methods are effective under nominal sensing conditions, they often rely on assumptions such as stable appearance, reliable correspondences, and stationary noise. In the presence of sensor degradation and adverse environmental conditions, these assumptions are frequently violated; different modalities may fail in distinct ways, which ultimately limits fusion robustness.
Motivated by these challenges, this paper introduces a robust, end-to-end deep learning–based multi-sensor fusion approach that adapts to environmental changes and sensor degradation (Figure~\ref{fig:motivation}). The proposed model is first trained on a simulated dataset and then fine-tuned on real-world data, demonstrating efficient simulation-to-reality transfer.
The main contributions of this paper are as follows:

\begin{itemize}
    \item We propose a novel end-to-end deep multi-sensor fusion framework that dynamically re-weights sensor contributions under diverse environmental conditions. By leveraging complementary multi-modal cues, the framework produces robust ego-pose estimates in challenging scenarios.

    \item We introduce an Adaptive Sensor Mixture-of-Experts (ASMoE) module with content-aware routing that assigns sensor weights at each timestep. In addition, decoupled translation and rotation branches preserve task-specific specialization while enabling cross-task information sharing via a cross-attention mechanism.

    \item We evaluate the proposed model under multiple sensor configurations, including a monocular camera, LiDAR, IMU, and radar, and demonstrate effective simulation-to-real transfer from a simulated environment to real-world data.
\end{itemize}

\section{RELATED WORK}

\subsection{Classical Methods}
Classical ego-pose estimation~\cite{thrun2000probabilistic} is typically formulated as geometric motion estimation from sensor measurements, coupled with probabilistic state estimation to model noise and uncertainty. In visual SLAM, feature-based pipelines such as ORB-SLAM2~\cite{mur2017orb} track sparse keypoints and refine poses via bundle adjustment, whereas direct methods such as LSD-SLAM~\cite{engel2014lsdslam} estimate motion through photometric alignment without explicit feature extraction. More recently, hybrid systems combine learned components with geometric optimization to improve robustness under challenging conditions, as demonstrated by H-SLAM~\cite{younes2024h}.
For visual--inertial odometry, classical approaches are broadly divided into filter-based methods such as MSCKF~\cite{mourikis2007msckf}, with consistency refinements to address linearization effects~\cite{li2013ekf}, and optimization-based formulations that solve nonlinear least-squares problems over visual--inertial constraints~\cite{leutenegger2013keyframe}. While these pipelines can be highly accurate under nominal conditions, performance often degrades in weakly constrained settings, where degeneracy and poor observability lead to drift.
% ~\cite{zhang2016degeneracy}

\subsection{Deep Learning-based Methods}
Driven by recent advances in neural architectures and computational resources, learning-based odometry, particularly when combined with multi-sensor fusion, has emerged as a viable alternative to traditional model-driven pipelines. Deep models can learn cross-modal associations directly from heterogeneous sensor streams, motivating end-to-end and adaptive fusion designs for autonomous driving scenarios in which sensor reliability varies across time and conditions~\cite{wei2025integrating}\cite{tang2023comparative}. Deep fusion strategies are commonly categorized by the stage at which fusion occurs: data-level fusion, which combines raw measurements early; feature-level fusion, which merges learned representations at an intermediate stage; and decision-level fusion, which fuses modality-specific predictions late. Feature-level fusion is often preferred when modalities degrade asymmetrically.

Uni-modal deep odometry methods also exhibit modality-specific failure modes. For instance, vision-based DeepVO~\cite{wang2017deepvo} is sensitive to poor texture and illumination; LiDAR-based ELiOT~\cite{lee2023eliot} may face generalization gaps. Consequently, bimodal systems such as DeepVIO~\cite{han2019deepvioselfsuperviseddeeplearning} and DEEPLIO~\cite{javanmardghareshiran2021deeplio} aim to exploit complementary modalities, but can remain vulnerable when the primary modality undergoes severe degradation. More recently, tri-modal approaches have targeted adverse conditions via adaptive reweighting and degradation-aware filtering, as exemplified by A2DO~\cite{lai2025a2do}.

Overall, prior work underscores the need for fusion strategies that preserve geometric coherence while incorporating content-adaptive sensor weighting to enable reliable ego-motion estimation under heterogeneous sensor noise and diverse operational conditions.

\section{METHODOLOGY}

\begin{figure*}[t]
    \centering
    \includegraphics[width=\textwidth]{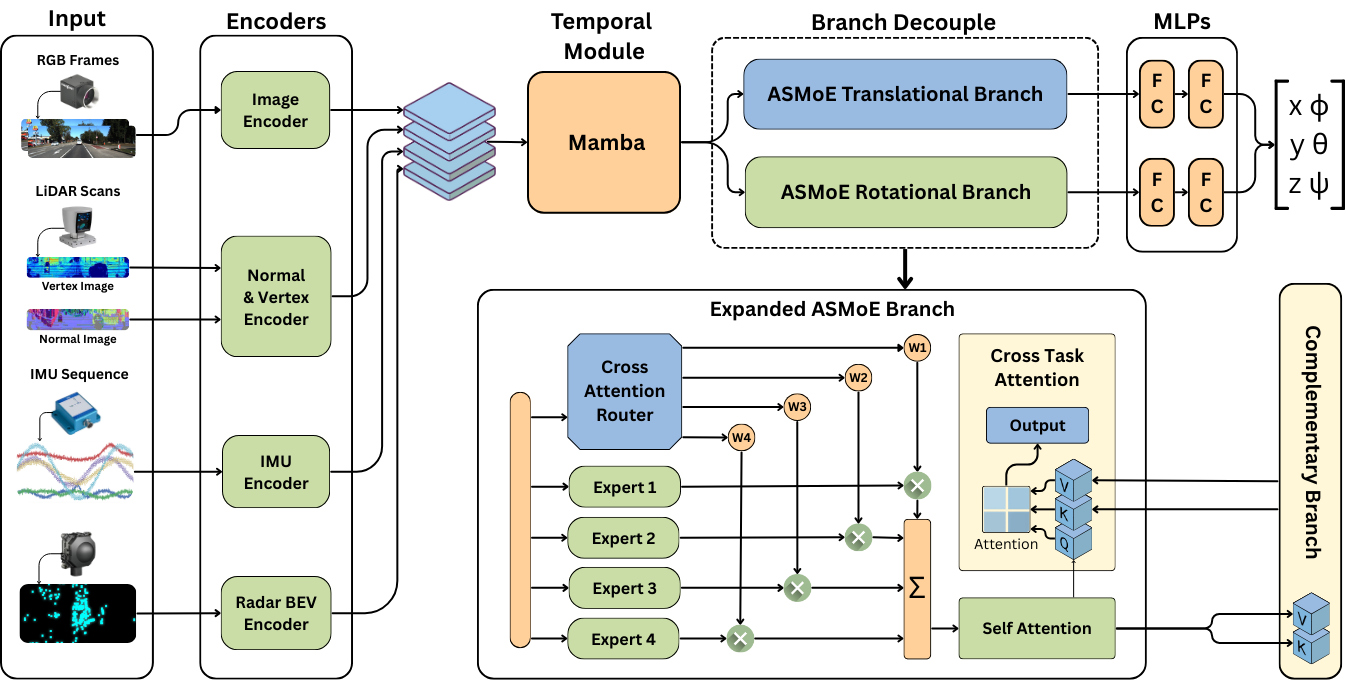}
    \caption{Overview of the proposed ADM-Fusion architecture. Multi-modal inputs (RGB, LiDAR vertex/normal images, IMU, and radar BEV) are encoded, temporally refined using per-sensor Mamba modules, and fused via an Adaptive Sensor Mixture-of-Experts (ASMoE) with content-aware routing. Translation and rotation are decoupled into separate fusion branches with cross-task attention, followed by temporal self-attention and MLP heads for 6-DoF pose prediction.}
    \label{fig:arch}
\end{figure*}

ADM-Fusion is an end-to-end, deep learning–based model with an encoder–decoder style architecture, as shown in Figure~\ref{fig:arch}. Each sensor’s raw measurements are processed by a modality-specific feature encoder, and the resulting feature representations are then passed through a temporal module to capture sequence-level dependencies. These temporally refined features are subsequently routed into two decoupled branches (one for translation and one for rotation), before being forwarded to the prediction head. This section details the individual components of the proposed architecture.

\subsection{Data Pre-processing}
Each training sequence is extracted using a sliding window of length $T$ with stride $r$. Each sequence contains the following inputs:
\begin{itemize}
    \item \textbf{RGB:} camera images $\mathbf{I} \in \mathbb{R}^{T \times C \times H \times W}$, which serve as the reference stream for temporal alignment of the remaining sensors.
    \item \textbf{IMU:} raw measurements between consecutive RGB frames. For each RGB timestep, we collect $N{=}16$ IMU samples, each represented as a 6D vector $[\boldsymbol{\omega}, \mathbf{a}]$ (gyroscope and accelerometer), forming $\mathbf{U} \in \mathbb{R}^{T \times N \times 6}$.
    \item \textbf{LiDAR:} point clouds are converted via spherical projection~\cite{milioto2019rangenet++} into two range-image representations: a \emph{vertex} image $\mathbf{L}_v$ encoding 3D coordinates $(x,y,z)$ and a \emph{normal} image $\mathbf{L}_n$ encoding surface normal components $(n_x,n_y,n_z)$. The LiDAR tensor is formed by channel-wise stacking, $\mathbf{L} = [\mathbf{L}_n;\mathbf{L}_v] \in \mathbb{R}^{T \times 6 \times H \times W}$.
    \item \textbf{Radar:} point clouds are projected into bird's-eye-view (BEV) images~\cite{liang2022bevfusion}, forming $\mathbf{R} \in \mathbb{R}^{T \times C \times H \times W}$, where the third channel encodes Doppler velocity.
\end{itemize}

\subsection{Feature Encoders}

\subsubsection{IMU Encoder}
IMU measurements are processed through an encoder consisting of two-layer GRU network \cite{dey2017gate} followed by a linear projection layer to map the features to a $d$-dimensional feature map.

\subsubsection{Image Encoder}
a ResNet backbone~\cite{he2016deep}, initialized with ImageNet-pretrained weights, is used to extract visual features from consecutive RGB frame pairs. This frame-pair stacking strategy is commonly adopted to capture short-term motion cues~\cite{wang2017deepvo}. For each timestep $t$, an input pair is formed by concatenating the previous and current frames along the channel dimension, $[\mathbf{I}_{t-1}; \mathbf{I}_{t}] \in \mathbb{R}^{6 \times H \times W}$. To accommodate the 6-channel input, the first convolutional layer is replaced with a 6-channel layer initialized by duplicating the pretrained 3-channel weights. The resulting features are processed by the standard ResNet layers, and a lightweight linear projection produces a $d$-dimensional representation. This encoder is fine-tuned end-to-end within the full architecture.

\subsubsection{LiDAR Encoder}
at each timestep, the LiDAR stacked tensor $\mathbf{L}$, composed of the normal image $\mathbf{L}_n$ and vertex image $\mathbf{L}_v$, is used to extract motion-aware features from a frame pair by comparing the projections at $t$ and $t-1$. Two lightweight CNN streams separately encode $\mathbf{L}_n$ and $\mathbf{L}_v$, after which their outputs are fused via channel concatenation. To capture inter-frame change, the previous ($t-1$) and current ($t$) feature maps, together with their difference, are then fed into a lightweight motion encoder. Finally, global average pooling followed by an MLP projection is applied to the resulting feature map.
%This approach uses feature differencing to model motion, demonstrating particular effectiveness for range-image LiDAR representations.%

\subsubsection{Radar BEV Encoder}
radar features are extracted from BEV representations using a RAFT-inspired~\cite{eslami2024rethinking} residual CNN encoder. For each timestep $t$, the previous and current BEV frames are concatenated along the channel dimension to form a frame-pair stack. The stacked pair is processed by a $7{\times}7$ convolution followed by three stages of ResNet-style residual blocks. Similar to the other modalities, a final linear projection maps the radar feature map into a $d$-dimensional space.

% To the best of our knowledge, this is the first end-to-end deep-learning based odometry model that fuse RGB, LiDAR, IMU, and radar in a single learned framework.

\subsection{Temporal Module}

Mamba is a selective state space model (SSM) for sequence modeling~\cite{gu2024mamba}. Its linear-time processing enables long-context temporal integration, which is useful in odometry to reduce drift and disambiguate locally uncertain motion. We apply temporal modeling after per-sensor feature extraction and before fusion so each modality is temporally refined independently prior to cross-modal weighting.

Let the per-sensor embeddings be stacked as $\mathbf{F}\in\mathbb{R}^{B\times T\times S\times D}$, where $B$ is batch size, $T$ sequence length, $S$ the number of sensors, and $D$ feature dimension. A Mamba module is applied along the time axis for each sensor stream, yielding temporally refined features that maintain per-sensor separation.

\subsection{Task-Decoupled Fusion Branches}
Temporal features from all available modalities are routed into two parallel branches: one for translation and one for rotation. This decoupling is motivated by the observation that different sensors do not contribute equally to all aspects of motion. IMU gyroscopes are generally reliable for capturing short-term angular dynamics, but their utility for translation is limited. In contrast, translation estimation is more susceptible to accelerometer integration drift and therefore relies more heavily on perceptual sensors such as camera images, LiDAR, and radar. By separating the branches, each path can learn specialized routing logic and task-specific feature transformations.

However, translation and rotation are not strictly independent: estimation errors in one component can systematically induce errors in the other. For this reason, cross-branch interaction is necessary at some stage.

Accordingly, each branch consists of: (i) an Adaptive Sensor Mixture-of-Experts (ASMoE) module for content-adaptive sensor weighting, followed by (ii) temporal self-attention for post-fusion temporal refinement, and (iii) cross-task attention to exchange complementary information between the translation and rotation branches.

Let the per-timestep multi-sensor feature tensor be
\(\mathbf{F}\in\mathbb{R}^{B\times T\times S\times D}\),
where \(B\) is the batch size, \(T\) is the sequence length, \(S\) is the number of sensors, and \(D\) is the feature dimension. The fusion module is applied independently at each timestep by reshaping \(\mathbf{F}\) into \(\tilde{\mathbf{F}}\in\mathbb{R}^{N\times S\times D}\), with \(N = B \cdot T\)

\subsubsection{Adaptive Sensor Mixture-of-Experts (ASMoE)}

ASMoE performs adaptive fusion over \(S\) sensors using two components: a set of per-sensor experts and an attention-based router. Given \(\tilde{\mathbf{F}}=\{\mathbf{x}_s\}_{s=1}^S\) with \(\mathbf{x}_s\in\mathbb{R}^{N\times D}\), each expert produces a task-specialized transformation:
\begin{equation}
\mathbf{e}_s = \mathrm{Expert}_s(\mathbf{x}_s)
= \mathrm{LN}\!\big(\mathbf{x}_s + \mathrm{MLP}(\mathbf{x}_s)\big),
\end{equation}
where \(\mathrm{MLP}(\cdot)\) is a two-layer feed-forward network with GELU activation, and \(\mathrm{LN}(\cdot)\) denotes LayerNorm. The residual connection preserves the original modality features while learning a refinement, allowing each expert to adapt the representation to the corresponding branch (translation or rotation) without over-distorting the input.

\subsubsection{Sensor Router}
to determine sensor weights, the router performs cross-sensor attention to explicitly model inter-modal relations. Given the sensor token set \(\tilde{\mathbf{F}}=\{\mathbf{x}_s\}_{s=1}^S\) with \(\mathbf{x}_s\in\mathbb{R}^{N\times D}\), we stack sensor tokens into \(\mathbf{X}=\mathrm{Stack}(\tilde{\mathbf{F}})\in\mathbb{R}^{S\times N\times D}\). The router then applies multi-head attention across the sensor dimension to obtain context-enhanced sensor representations:
\begin{align}
% \mathbf{Q} &= \mathbf{X}\mathbf{W}_Q,\qquad 
% \mathbf{K} = \mathbf{X}\mathbf{W}_K,\qquad 
% \mathbf{V} = \mathbf{X}\mathbf{W}_V, \\
\mathbf{A} &= \mathrm{softmax}\!\left(\frac{\mathbf{Q}\mathbf{K}^\top}{\sqrt{d_h}}\right), \\
\mathbf{y}_s &=\sum_{j=1}^S \mathbf{A}_{s,j}\mathbf{V}_j,
\end{align}
where \(\mathbf{Q}\), \(\mathbf{K}\), and \(\mathbf{V}\) are the \emph{query}, \emph{key}, and \emph{value} projections obtained via learned linear maps~\cite{vaswani2017attention}. Here, \(\mathbf{A}_{s,j}\) denotes the attention assigned from sensor \(s\) to sensor \(j\) (per token), and \(\mathbf{V}_j\) denotes the value projection corresponding to sensor \(j\).

A lightweight MLP maps each attended sensor output \(\mathbf{y}_s\) to a scalar routing logit:
\begin{equation}
\ell_s = \mathrm{MLP}_{\text{route}}(\mathbf{y}_s).
\end{equation}
Sensor weights are then obtained via a softmax over sensors:
\begin{equation}
w_s = \frac{\exp(\ell_s)}{\sum_{j=1}^S \exp(\ell_j)}.
\end{equation}

Finally, ASMoE fuses the expert outputs using the routing weights:
\begin{equation}
\mathbf{z} = \sum_{s=1}^S w_s\,\mathbf{e}_s,
\end{equation}
followed by an output projection.

\subsubsection{Temporal \& Cross-Modality Attention}
let \(\mathbf{z}^{t}, \mathbf{z}^{r} \in \mathbb{R}^{B\times T\times D}\) denote the fused sequences output by ASMoE for the translation and rotation branches, respectively. We first refine each branch temporally using self-attention, and then enable controlled interaction using cross-attention.

\paragraph{Temporal self-attention} is applied independently to each branch:
\begin{equation}
    \hat{\mathbf{z}}^{\{t,r\}} = \mathrm{MHA}\!\left(\mathbf{z}^{\{t,r\}}\right),
\end{equation}
where \(\mathrm{MHA}(\cdot)\) denotes a standard multi-head self-attention block operating over the \(T\) temporal positions.

\paragraph{Cross-Task Attention}
although translation and rotation are estimated through separate branches, these processes are not independent, as systematic errors in one can propagate to the other. To facilitate controlled information exchange while maintaining the distinction between pathways, cross-task attention is applied after temporal refinement. In this formulation, each branch queries the representation learned by the other, enabling it to correct or contextualize its own estimate using complementary motion cues:
\begin{align}
    \mathbf{h}^t &= \mathrm{CrossAttn}\!\left(\mathbf{Q}=\hat{\mathbf{z}}^t,\;\mathbf{K}=\hat{\mathbf{z}}^r,\;\mathbf{V}=\hat{\mathbf{z}}^r\right), \\
    \mathbf{h}^r &= \mathrm{CrossAttn}\!\left(\mathbf{Q}=\hat{\mathbf{z}}^r,\;\mathbf{K}=\hat{\mathbf{z}}^t,\;\mathbf{V}=\hat{\mathbf{z}}^t\right),
\end{align}
where \(\mathbf{h}^t, \mathbf{h}^r \in \mathbb{R}^{B \times T \times D}\) are the final branch representations passed to the pose regression heads.

\subsection{Pose Regression Head}
The refined branch representations \(\mathbf{h}^t\) and \(\mathbf{h}^r\) are passed to dedicated regression heads, each consisting of a LayerNorm followed by a two-layer MLP with GELU activation. The translation head predicts a three-dimensional displacement vector \((x, y, z)\), while the rotation head predicts the corresponding Euler angles \((\phi, \theta, \psi)\), together forming a complete per-frame 6-DoF pose estimate.

\subsection{Loss Function}

We supervise relative pose using frame-to-frame (F2F) losses on relative motion and
frame-to-global (F2G) losses on the accumulated trajectory.
Let $\hat{\mathbf{t}}_i$ and $\hat{\boldsymbol{\theta}}_i$ denote predicted relative
translation and Euler angles at timestep $i$, with ground truth
$\mathbf{t}_i$ and $\boldsymbol{\theta}_i$.

\paragraph{Translation and Rotation Loss}
the F2F losses are defined as
\begin{equation}
L_{\mathrm{F2F}}^t=\frac{1}{T}\sum_{i=1}^T \rho(\hat{\mathbf{t}}_i-\mathbf{t}_i),\quad
L_{\mathrm{F2F}}^r=\frac{1}{T}\sum_{i=1}^T
d(\mathbf{R}(\hat{\boldsymbol{\theta}}_i),\mathbf{R}(\boldsymbol{\theta}_i)),
\end{equation}
where $\rho(\cdot)$ is the Huber loss and $d(\cdot)$ is the geodesic
distance on $\mathrm{SO}(3)$.

To enforce trajectory-level consistency, relative motions are accumulated
to global poses $(\hat{\mathbf{p}}_i,\hat{\mathbf{R}}_i)$ and
$(\mathbf{p}_i,\mathbf{R}_i)$:
\begin{equation}
L_{\mathrm{F2G}}^t=\frac{1}{T}\sum_{i=1}^T\rho(\hat{\mathbf{p}}_i-\mathbf{p}_i),\quad
L_{\mathrm{F2G}}^r=\frac{1}{T}\sum_{i=1}^T d(\hat{\mathbf{R}}_i,\mathbf{R}_i).
\end{equation}

\paragraph{Homoscedastic Weighting}
to balance translation and rotation scales we adopt the uncertainty weighting of~\cite{kendall2018multi}:
\begin{equation}
L_{\text{pose}} = e^{-s_t}(L^{t}_{\text{F2F}} + L^{t}_{\text{F2G}})
+ e^{-s_r}(L^{r}_{\text{F2F}} + L^{r}_{\text{F2G}})
+ s_t + s_r .
\end{equation}
where $s_t$ and $s_r$ are learnable log-variance parameters that automatically balance the translation and rotation loss terms during training.

\paragraph{Balancing Regularization}
to avoid router collapse in ASMoE we encourage uniform expert usage.
Let $f_s = \frac{1}{N}\sum_n w_{n,s}$ be the mean routing weight for expert $s$. The regularizer is
\begin{equation}
L_{\text{bal}} = \frac{1}{S}\sum_{s=1}^{S}\left(f_s-\frac{1}{S}\right)^2 .
\end{equation}

\paragraph{Total Loss}
\begin{equation}
L_{\text{total}} = L_{\text{pose}} + \lambda_{\text{bal}} L_{\text{bal}} .
\end{equation}
where $\lambda_{\text{bal}}$ is a tunable coefficient controlling the strength of the balancing regularization. In practice, $\lambda_{\text{bal}}$ is selected through a small hyperparameter sweep based on validation ATE/RPE performance.
% \vspace{5mm}
\section{EXPERIMENTS}

\subsection{Experimental Setups}
\subsubsection{Dataset}
We pretrain on the simulated CARLA-Loc dataset~\cite{han2023carla} and fine-tune on KITTI Odometry~\cite{geiger2013vision}.
\begin{itemize}[leftmargin=1.0em, labelsep=0.6em, itemindent=0pt, listparindent=0pt]
 \item \textbf{CARLA-Loc:} a multi-sensor simulated odometry dataset spanning seven towns with clear, rainy, and foggy weather in static and dynamic traffic. It provides synchronized RGB, LiDAR, IMU, and radar. We train on all towns except Town05, which is used for testing across its six scenarios.

 \item \textbf{KITTI Odometry:} a real-world driving benchmark with camera, LiDAR, and OXTS ground truth. We train on sequences $\{00,01,02,04,06,08,09\}$ and evaluate on $\{05,07,10\}$.
\end{itemize}

\subsubsection{Evaluation Metrics}
we evaluate odometry performance using the \texttt{evo} toolkit~\cite{grupp2017evo} and report: (i) raw and ground truth aligned Absolute Trajectory Error (ATE) RMSE in meters (m); (ii) Relative Pose Error (RPE) for translation in meters (m) and rotation in degrees (\(^{\circ}\)); and (iii) the KITTI odometry metrics, namely the mean translational error \(t_{\mathrm{rel}}\) (\%) and rotational error \(r_{\mathrm{rel}}\) (deg/100\,m), averaged over 100\,m path segments.

\subsubsection{Implementation Details}
All experiments are conducted on a single NVIDIA RTX 4090 GPU. We train the fusion model using the AdamW optimizer with an initial learning rate of \(10^{-3}\), a batch size of 4, and a sequence length of 10. In the full four-sensor configuration, ADM-Fusion contains \(34\)M trainable parameters. The model is pretrained on the CARLA-Loc simulated dataset for 50 epochs and then fine-tuned on the KITTI Odometry benchmark for 50 epochs. For runtime evaluation, inference on an NVIDIA RTX 3050 Ti achieves real-time performance of 60--70 frames per second (FPS).

\begin{figure*}[t]
\centering
\includegraphics[width=\textwidth]{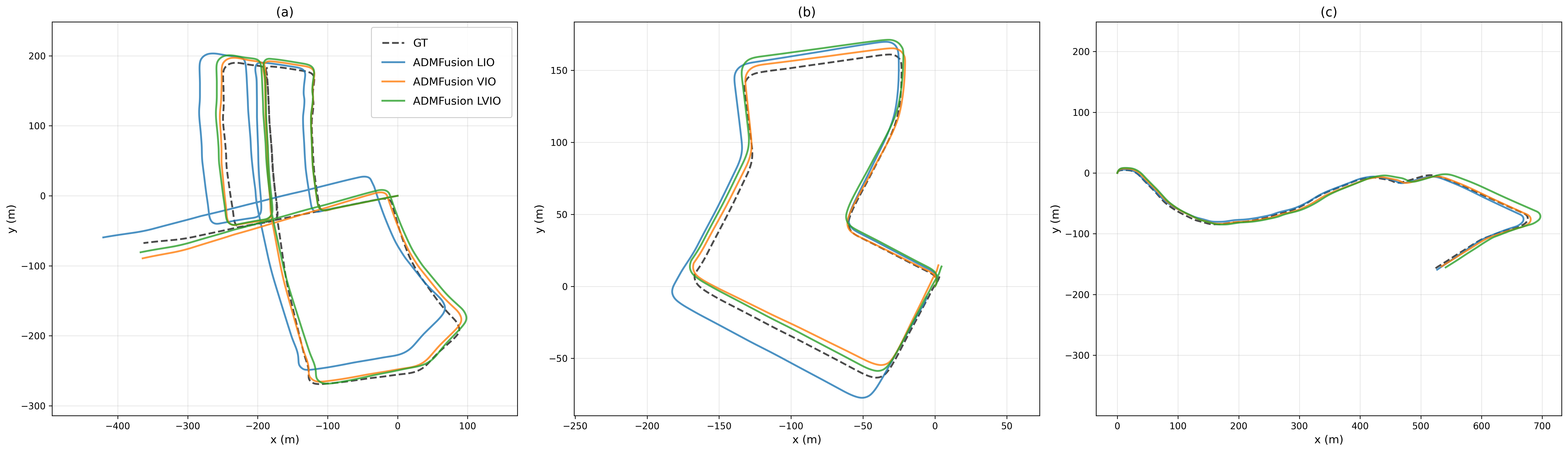}
\caption{\footnotesize Qualitative trajectory comparison on KITTI Odometry sequences 05, 07, and 10 (subplots (a)--(c)).
GT denotes ground truth. ADM-Fusion variants correspond to LiDAR--IMU (LIO), Vision--IMU (VIO), and LiDAR--Vision--IMU (LVIO).}
\label{fig:kitti_traj_comp}
\end{figure*}
\subsection{Discussion}
\label{sec:discussion_kitti}

\begin{figure}[t]
\centering
\includegraphics[width=\columnwidth]{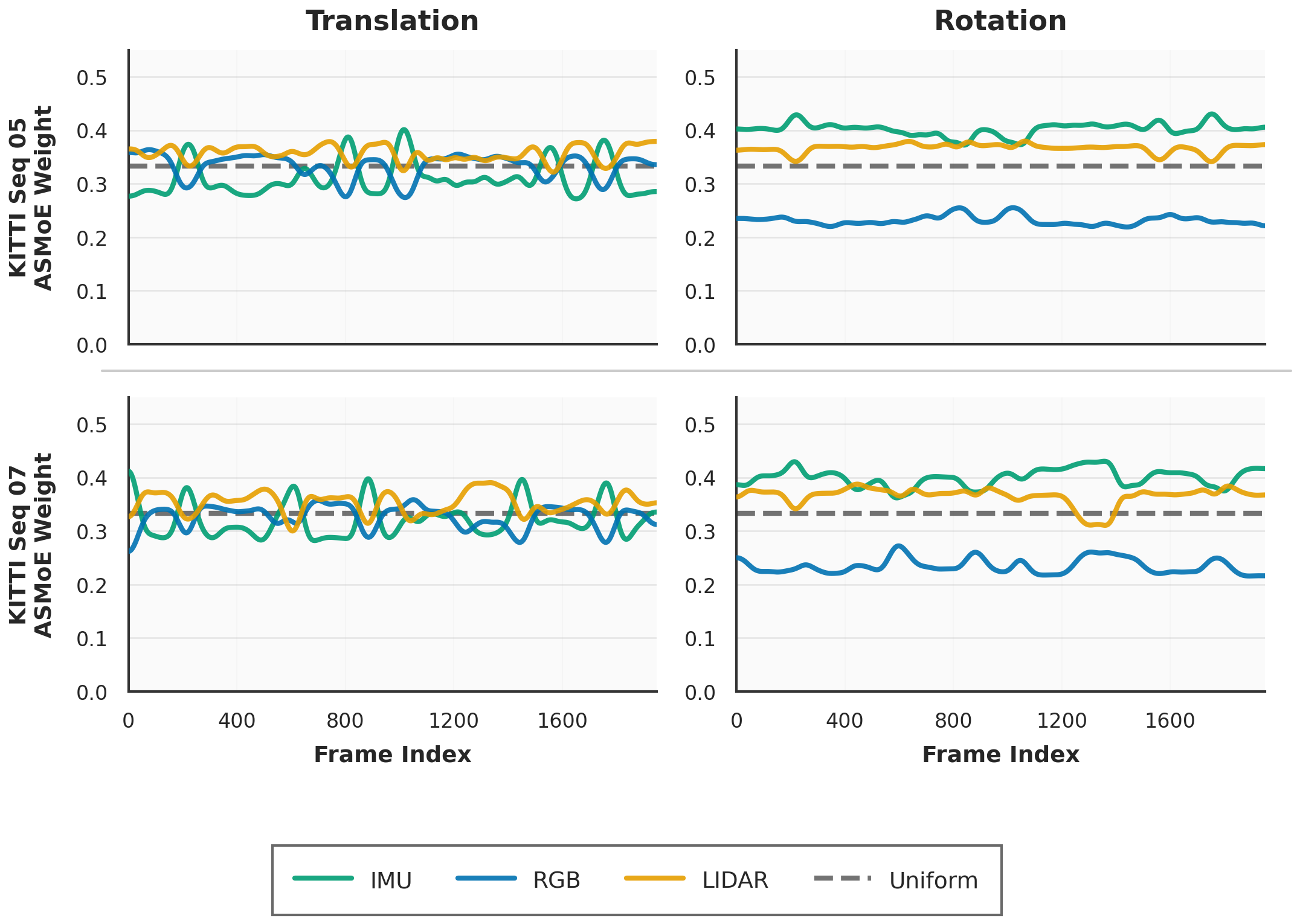}
\caption{Inference ASMoE modality weights for translation and rotation on KITTI sequences 05 and 07 using LVIO configuration.}
\label{fig:asmoe_weights}
\end{figure}

\subsubsection{Real World Evaluation}
the KITTI Odometry benchmark provides a real-world evaluation setting characterized by complex vehicle dynamics, extended trajectories, and substantial accumulated drift. Table~\ref{tab:kitti_grouped} presents the quantitative performance of ADM-Fusion under LiDAR--IMU (LIO), Vision--IMU (VIO), and LiDAR--Vision--IMU (LVIO) configurations on sequences 05, 07, and 10.

The most consistent gains are observed in rotational accuracy: the three-sensor LVIO configuration attains the lowest mean rotational drift, indicating that complementary modalities are particularly effective for constraining heading over long trajectories. In contrast, translational drift does not improve as consistently, despite the improved orientation estimates. This suggests that translation is more sensitive to residual scale or velocity biases, calibration errors, and long-term accumulation effects. This modality-dependent behavior supports the use of separate, task-specific branches, as LiDAR, vision, and IMU each provide distinct information and exhibit different noise characteristics. Dedicated branches can preserve modality-specialized features prior to fusion, rather than forcing a single shared representation to address all error modes equally. The qualitative trajectories in Fig.~\ref{fig:kitti_traj_comp} further reflect this trend: LVIO follows the ground truth more closely during turns and extended segments, whereas two-sensor configurations accumulate orientation errors that lead to observable divergence.

Figure~\ref{fig:asmoe_weights} presents the real-time ASMoE modality weights produced by ADM-Fusion during inference on KITTI sequences 05 and 07 under the LVIO configuration. Two primary trends are observed. First, the weights display temporal smoothness while remaining adaptive; rather than favoring a single sensor, the model continuously adjusts modality contributions as the motion behavior changes. Second, sensor importance varies according to the task. The IMU generally receives higher weighting in the rotation branch, reflecting its capacity to capture short-term angular dynamics, whereas LiDAR and RGB contribute more substantially to translation stabilization. The weighting remains well-distributed over time, indicating no persistent sensor dominance or collapse. This result is attributed to the balance loss, which regularizes the router and encourages sustained utilization of all modalities. Collectively, these patterns suggest that ASMoE learns dynamic, context-dependent fusion strategies instead of relying on fixed rules.

\begin{table}[t]
\centering
\scriptsize
\setlength{\tabcolsep}{1.8pt}
\renewcommand{\arraystretch}{1.08}
\caption{KITTI drift metrics ($t_{\mathrm{rel}}$, $r_{\mathrm{rel}}$) on sequences 05/07/10 and their mean.}
\begin{tabular}{lcccccccc}
\toprule
\textbf{Experiments} &
\multicolumn{2}{c}{\textbf{05}} &
\multicolumn{2}{c}{\textbf{07}} &
\multicolumn{2}{c}{\textbf{10}} &
\multicolumn{2}{c}{\textbf{overall}} \\
\cmidrule(lr){2-3}\cmidrule(lr){4-5}\cmidrule(lr){6-7}\cmidrule(lr){8-9}
& $t_{\text{rel}}(\%)$ & $r(^{\circ})$
& $t_{\text{rel}}(\%)$ & $r(^{\circ})$
& $t_{\text{rel}}(\%)$ & $r(^{\circ})$
& $t_{\text{rel}}(\%)$ & $r(^{\circ})$ \\
\midrule

\multicolumn{9}{c}{\textbf{LiDAR--IMU}} \\
\midrule

LIO-SAM~\cite{shan2020lio}
& 1.69 & 1.28 
& 2.87 & 1.62 
& 4.97 & 2.17 
& 3.18 & 1.69 \\

DEEPLIO~\cite{javanmardghareshiran2021deeplio}     
& \textbf{1.24} & \textbf{0.21}  
& \textbf{1.92} & \textbf{0.32}  
& 4.0 & 0.51  
& \textbf{2.38} & \textbf{0.35}  \\

A2DO~\cite{lai2025a2do}
& 3.84 & 1.85 
& 3.21 & 2.51 
& 4.80 & 1.69 
& 3.95 & 2.02 \\

\textbf{ADM-Fusion}
& 3.15 & 0.4   
& 3.92 &  0.36
& \textbf{2.21} & \textbf{0.49} 
& 3.09 & 0.44  \\
\midrule

\multicolumn{9}{c}{\textbf{Vision--IMU}} \\
\midrule
% VINS-Mono~\cite{qin2018vins}     
% & 11.6 & 1.26 
% & 10.0 & 1.72 
% & 16.5 & 2.34 
% & 12.7 & 1.77 \\

ATVIO~\cite{liu2021atvio}         
& 4.93 & 2.40 
& 3.78 & 2.59 
& 5.71 & 2.96 
& 4.81 & 2.65 \\

SelectFusion(Hard)~\cite{9788038}
& 4.11 & 1.49  
& 3.44  & 1.86  
& \textbf{1.51} & 0.91 
& 3.02 & 1.42  \\

A2DO
& 2.95 & 1.40 
& 3.98 & 2.90 
& 4.36 & 1.52 
& 3.76 & 1.94 \\

\textbf{ADM-Fusion}
& \textbf{1.96}  & \textbf{0.55}  
& \textbf{2.07} & \textbf{0.81} 
& 2.19 & \textbf{0.38} 
& \textbf{2.07} & \textbf{0.58}   \\
\midrule

\multicolumn{9}{c}{\textbf{LiDAR--Vision--IMU}} \\
\midrule

% A2DO (w/o pre-training)
% & 2.93 & 0.76 
% & 3.30 & 1.19 
% & 3.29 & 0.90 
% & 3.17 & 0.95 \\

A2DO
& \textbf{1.24} & 0.44
& \textbf{1.07} & 0.67
& \textbf{1.77} & 0.50
& \textbf{1.36} & 0.54 \\

% \textbf{ADMFusion} (w/o pre-training)
% & - & -   
% & - & -  
% & - & -  
% & - & -  \\

\textbf{ADM-Fusion}
& 2.30 & \textbf{0.41}   
& 1.93 & \textbf{0.28}  
& 2.87 & \textbf{0.32}  
& 2.36 & \textbf{0.33}  \\

\bottomrule
\end{tabular}
\label{tab:kitti_grouped}
\end{table}
% Pretraining Table : Carla Loc (Simulated)

Compared with the state-of-the-art baselines in Table~\ref{tab:kitti_grouped}, ADM-Fusion shows competitive performance across sensing configurations. In the two-sensor settings (LIO and VIO), ADM-Fusion consistently yields lower drift, improving both translational and rotational accuracy across the evaluated sequences. These results suggest that the proposed adaptive fusion more effectively balances modality contributions under partial-modality conditions.

In the three-sensor LVIO configuration, ADM-Fusion achieves lower mean rotational drift than A2DO, indicating improved heading stability over extended trajectories. In contrast, A2DO attains a marginally lower mean translational drift in this setting. This divergence suggests that while ADM-Fusion is particularly effective at constraining orientation through complementary modality interaction, translational accuracy remains influenced by velocity bias, scale consistency, and long-term accumulation effects. Overall, ADM-Fusion provides the most consistent rotational stability across configurations while maintaining competitive translational accuracy.

\subsubsection{CARLA-Loc Synthetic Evaluation}
Tables~\ref{tab:lio_imu_ate}--\ref{tab:carla_map05_ate} report Absolute Trajectory Error (ATE) on CARLA-Loc under different weather and motion conditions. We report both raw ATE and SE(3)-aligned ATE (marked *): raw ATE reflects absolute drift, while aligned ATE removes a single best-fit rigid transform to highlight relative trajectory consistency. In the LIO configuration (Table~\ref{tab:lio_imu_ate}), ADM-Fusion demonstrates competitive performance but does not surpass highly optimized classical pipelines such as FASTLIO2~\cite{xu2022fast}. Notably, when transitioning from static to dynamic environments, the model remains robust: ATE increases by only 0.28\,m, indicating resilience to increased scene complexity.

In VIO settings, ADM-Fusion achieves state-of-the-art performance among the reported VIO baselines in both static and dynamic scenarios, demonstrating robustness to appearance degradation and motion. Across backbone architectures, ResNet34 yields the best overall average, while ResNet18 performs better at night, suggesting improved invariance under challenging visual conditions.

In the three-sensor LVIO configuration, A2DO achieves a lower average ATE (0.82 m) than ADM-Fusion on the synthetic CARLA benchmark (2.69 m). This performance gap highlights a key property of content-aware routing: when sensor inputs are uniformly reliable, as is typical for LiDAR in clean simulation environments, the routing mechanism lacks sufficient quality contrast across modalities to effectively differentiate sensor contributions. Under these conditions, A2DO's design remains effective, while the adaptive routing in ASMoE provides limited benefit. Supporting this interpretation, integrating LiDAR into VIO does not consistently enhance performance in CARLA, whereas augmenting LIO with RGB generally improves stability under dynamic conditions. This finding suggests that visual features provide complementary rather than redundant information even in synthetic scenarios. On the KITTI dataset, where real-world LiDAR is affected by motion distortion and noise, the routing mechanism becomes more effective: LVIO achieves the lowest rotational drift across all configurations (Table~\ref{tab:kitti_grouped}), indicating that ASMoE's adaptive weighting is most advantageous when sensor reliability varies over time and across conditions.
\begin{table}[tp]
\centering
\scriptsize
\setlength{\tabcolsep}{3.2pt}
\renewcommand{\arraystretch}{0.85}
\caption{\footnotesize ATE(m) on CARLA-Loc Map05 using LiDAR--IMU configuration. * denotes aligned ATE}
\begin{tabular}{lccc}
\toprule
\textbf{Method} & \textbf{Static} & \textbf{Dynamic} & \textbf{Avg.} \\
\midrule
ALOAM~\cite{zhang2014loam}      & 4.53  & 93.64 & 49.08  \\
FASTLIO2~\cite{xu2022fast}     & \textbf{2.36}  & \textbf{2.70} & \textbf{2.53} \\
A2DO     & 2.88  & 4.06 & 3.47 \\
ADM-Fusion    & 4.67 & 4.95 & 4.81 \\
ADM-Fusion*  & 4.08 & 3.22 & 3.65 \\
\bottomrule
\end{tabular}

\label{tab:lio_imu_ate}
\end{table}
\begin{table}[tp]
\centering
\scriptsize
\setlength{\tabcolsep}{2.2pt}
\renewcommand{\arraystretch}{0.85}
\caption{\footnotesize ATE (m) on CARLA-Loc Town05.  
CN=Clear Noon, FN=Foggy Noon, RN=Rainy Night; S=Static, D=Dynamic. * denotes aligned ATE}
\resizebox{\linewidth}{!}{%
\begin{tabular}{l ccccccc}
\toprule
\textbf{Method} & \textbf{CN(S)} & \textbf{FN(S)} & \textbf{RN(S)} & \textbf{CN(D)} & \textbf{FN(D)} & \textbf{RN(D)} & \textbf{Avg.} \\
\midrule
\multicolumn{8}{c}{\textbf{Vision--IMU}} \\
\midrule
ORB3-SVIO~\cite{campos2021orb}        & 3.24 & 23.52 & 18.03 & 2.29 & 555.48 & 425.74 & 171.38 \\
VINS-SVIO~\cite{qin2025general}        & 4.03 & \textit{fail} & \textit{fail} & 3.97 & \textit{fail} & 6.76 & -- \\
A2DO(ResNet34)          & 2.23 & 2.21 & 4.42 & 3.04 & 1.96 & 3.55 & 2.90 \\
ADM-Fusion(ResNet18)    & 3.21 & 3.91 & \textbf{3.24} & 2.84 & 4.44 & \textbf{3.21} & 3.48 \\
ADM-Fusion(ResNet34)    & 2.02 & 1.76 & 3.98 & 1.46 & 1.69 & 3.75 & 2.44 \\
ADM-Fusion(ResNet34)*   & \textbf{1.34} & \textbf{1.41} & \textbf{2.42} & \textbf{1.09} & \textbf{1.27} & \textbf{2.19} & \textbf{1.62} \\
\midrule

\multicolumn{8}{c}{\textbf{LiDAR--Vision--IMU}} \\
\midrule
A2DO        & \textbf{0.34} & \textbf{0.34} & \textbf{0.65} & \textbf{0.94} & \textbf{0.77} & \textbf{1.91} & \textbf{0.82} \\
ADM-Fusion  & 2.49 & 2.13 & 2.77 & 3.55 & 2.21 & 3.03 & 2.69 \\
ADM-Fusion*   & 1.45 & 1.58 & 2.54 & 1.19 & 1.32 & 2.58 & 1.77  \\
\bottomrule
\end{tabular}
}
\label{tab:carla_map05_ate}
\end{table}

% \vspace{3mm}
Integrating radar into the LVIO configuration significantly enhances translational consistency, reducing average translational RPE from 26.93 mm to 9.57 mm, representing a 64\% reduction, while causing only a marginal increase in rotational RPE (from 0.025° to 0.030°, Table~\ref{tab:rpe_town05_vlio_rvlio}). This asymmetric improvement highlights radar's complementary function: its Doppler velocity measurements directly constrain short-horizon translational drift, a capability not provided by camera or LiDAR, while contributing minimally to orientation estimation, which remains well-constrained by IMU and LiDAR geometry. The consistency of this improvement across all six Town05 conditions further demonstrates ADM-Fusion's capacity to effectively integrate a fourth modality and leverage sensor-specific strengths through the ASMoE routing mechanism.

\begin{table}[tp]
\centering
\scriptsize
\setlength{\tabcolsep}{3.2pt}
\renewcommand{\arraystretch}{0.95}
\caption{RPE RMSE on Town05 for ADM-Fusion configurations: 4-sensor RLVIO vs 3-sensor LVIO.}
\label{tab:rpe_town05_vlio_rvlio}
\begin{tabular}{llccccccc}
\toprule
\textbf{Config} & \textbf{Metric} &
\textbf{CN(S)} & \textbf{CN(D)} &
\textbf{FN(S)} & \textbf{FN(D)} &
\textbf{RN(S)} & \textbf{RN(D)} &
\textbf{Avg.} \\
\midrule

\multirow{2}{*}{\textbf{LVIO}} & RPE$_t$ (mm)  & 20.53 & 22.75 & 27.58 & 28.78 & 32.20 & 29.77 & 26.93 \\
                              & RPE$_r$ (deg) & 0.025 & 0.025 & 0.026 & 0.026 & 0.025 & 0.026 & 0.025 \\
\midrule
\multirow{2}{*}{\textbf{RLVIO}} & RPE$_t$ (mm)  & 9.15 & 9.56 & 9.29 & 9.67 & 9.91 & 9.87 & 9.57 \\
                               & RPE$_r$ (deg) & 0.029 & 0.029 & 0.031 & 0.030 & 0.030 & 0.029 & 0.030 \\
\bottomrule
\end{tabular}
\end{table}

\subsubsection{Ablation Study}
Table~\ref{tab:ablation_carla} analyzes the impact of the ASMoE routing strategy and temporal design in the LIO configuration on CARLA-Loc using KITTI-style drift metrics. The router is the core of the adaptive weighting mechanism, estimating per-sensor reliability weights from temporally enhanced features.

Among the evaluated routing strategies, Graph Attention Networks (GAT)~\cite{velivckovic2017graph} perform the worst. This can be attributed to the small number of modality nodes (at most four), for which the relational inductive bias of GAT is less beneficial. A simple multi-layer neural network (NN) router achieves competitive overall performance, particularly in translational drift. However, the per-sensor attention router yields stronger rotational accuracy and more consistent behavior across static and dynamic scenes. Weighting entire sensor feature maps is more effective than per-dimension weighting, highlighting the importance of preserving modality-level structure during fusion. Moreover, a hybrid design using per-sensor attention in the rotational branch and the NN router in the translational branch does not improve performance, suggesting that consistent routing mechanisms across branches are preferable. Finally, the uncertainty-based routing approach degrades performance in dynamic scenarios, likely due to instability in uncertainty estimation under increased motion complexity.

% Ablation Studies Table 
\begin{table}[tp]
\centering
\scriptsize
\setlength{\tabcolsep}{3.0pt}
\renewcommand{\arraystretch}{1.08}
\caption{Ablation study using KITTI-style drift metrics on CARLA-Loc Town05 using LiDAR-IMU configuration. w and w/o denote with and without the respective module.}
\begin{tabular}{lcccccc}
\toprule
\textbf{Experiments} &
\multicolumn{2}{c}{\textbf{Static}} &
\multicolumn{2}{c}{\textbf{Dynamic}} &
\multicolumn{2}{c}{\textbf{Overall}} \\
\cmidrule(lr){2-3}\cmidrule(lr){4-5}\cmidrule(lr){6-7}
& $t_{\text{rel}}$(\%) & $r$($^\circ$)
& $t_{\text{rel}}$(\%) & $r$($^\circ$)
& $t_{\text{rel}}$(\%) & $r$($^\circ$) \\
\midrule
\multicolumn{7}{c}{\textbf{ASMoE Router}} \\
\midrule
Graph Attention Network          & 6.28 & 1.77 & 6.21 & 1.82 & 6.25 & 1.79 \\
Per-Sensor Feature Attention     & 3.01 & 1.17 & 3.11 & 1.17 & 3.06 & 1.17 \\
Uncertainty-based                & 2.88 & 0.64 & 3.13 & 0.65 & 3.01 & 0.65 \\
NN w Attention                   & 2.76 & 0.77 & 2.67 & 0.81 & 2.72 & 0.79 \\
Neural Network (NN)              & \textbf{2.61} & 0.72 & \textbf{2.42} & 0.72 & \textbf{2.52} & 0.72 \\
Per-Sensor Attention             & 2.82 & \textbf{0.54} & 2.46 & \textbf{0.70} & 2.64 & \textbf{0.62} \\
\midrule 
\multicolumn{7}{c}{\textbf{Temporal Module}} \\
\midrule
Pre-Mamba w Post-GRU     & 3.47 & \textbf{0.72} & 3.43 & 0.75 & 3.45 & 0.74 \\
Pre-Mamba w/o Post-GRU     & \textbf{2.61} & \textbf{0.72} & \textbf{2.42} & \textbf{0.72} & \textbf{2.52} & \textbf{0.72} \\
Pre-GRU w/o Post-GRU       & 3.86   & 1.46   & 3.95   & 1.54   & 3.91 & 1.5 \\
\midrule
\multicolumn{7}{c}{\textbf{Balance Loss}} \\
\midrule
ADM-Fusion w/o Balance Loss       & 3.7 & 0.89 & 2.89 & 0.93 & 3.29 & 0.91 \\
ADM-Fusion w Balance Loss         & \textbf{2.82} & \textbf{0.54} & \textbf{2.46} & \textbf{0.70} & \textbf{2.64} & \textbf{0.62} \\
% \midrule
% \multicolumn{7}{c}{\textbf{Cross-Task}} \\
% ADM-Fusion w/o Cross-Task   & 2.11 & 0.47 & 2.20 &  0.50 & 2.16 & 0.485 \\
% ADM-Fusion w Cross-Task  & 2.82 & 0.54 & 2.46 & 0.70 & 2.64 & 0.62 \\

% \midrule
% ADM-Fusion w/o Cross Task       & 3.7 & 0.89 & 2.89 & 0.93 & 3.29 & 0.91 \\
% ADM-Fusion w Cross Task         & \textbf{2.82} & \textbf{0.54} & \textbf{2.46} & \textbf{0.70} & \textbf{2.64} & \textbf{0.62} \\
 
\bottomrule
\end{tabular}

\label{tab:ablation_carla}
\end{table}

For temporal modeling, pre-fusion temporal enhancement with Mamba consistently outperforms a GRU. This indicates that richer sequence modeling prior to fusion improves the reliability of per-sensor feature estimation. Adding a post-fusion temporal module does not yield further gains, suggesting that temporal reasoning is most effective when applied at the modality level before adaptive weighting.

The impact of balance loss regularization is further evaluated. Its removal results in a significant decline in both translational and rotational accuracy, as overall $t_{\text{rel}}$ increases from 2.64\% to 3.29\% and $r$ rises from 0.62\textdegree{} to 0.91\textdegree{}. These findings indicate that, in the absence of balance loss, the router exhibits sensor collapse by over-relying on a single dominant modality instead of maintaining distributed utilization across all sensors. Therefore, balance loss serves a crucial regularization function, ensuring that ASMoE learns robust and diverse sensor contributions rather than converging to suboptimal routing solutions.

Overall, the per-sensor attention router combined with pre-fusion Mamba provides the best trade-off between translational and rotational accuracy and improves robustness across varying conditions.

\section{Conclusion}
This paper proposes an end-to-end, deep learning–based multi-sensor fusion approach that adapts to environmental changes and sensor degradation. The model employs an Adaptive Sensor Mixture-of-Experts (ASMoE) to weigh each sensor according to degradation and noise conditions. In addition, translation and rotation are decoupled to preserve task-specific specialization, while cross-attention enables controlled information exchange to account for the coupling between the two motion components. Experimental results show that the proposed method remains robust across different weather conditions and dynamic scenes. The model also demonstrates strong sim-to-real transfer through pretraining on a simulated dataset and fine-tuning on real-world data, suggesting potential for real-time deployment.

Future work will extend ASMoE with calibrated uncertainty to enable reliability-aware sensor dropout under severe degradation, and explore pretrained foundation-model encoders to improve robustness to noise and domain shift.
%%%%%%%%%%%%%%%%%%%%%%%%%%%%%%%%%%%%%%%%%%%%%%%%%%%%%%%%%%%%%%%%%%%%%%%%%%%%%%%%

\bibliographystyle{IEEEtran}
\bibliography{references}

\end{document}